\documentclass[letterpaper, 10 pt, conference]{ieeeconf}  

\IEEEoverridecommandlockouts                              

\title{\LARGE \bf
Augmentation for Learning From Demonstration with Environmental Constraints}

\usepackage{graphics} 
\usepackage{epsfig} 
\usepackage{mathptmx} 
\usepackage{times} 
\usepackage[cmex10]{amsmath}
\usepackage{amssymb}  
\usepackage{ifpdf}
\usepackage{xcolor}
\definecolor{sd}{gray}{0.5}
\usepackage{colortbl}
\usepackage{subcaption}
\usepackage{graphicx}
\usepackage{balance}
\usepackage[algo2e]{algorithm2e} 
\usepackage{algorithm}
\usepackage[noend]{algorithmic}
\usepackage{framed}
\usepackage{tabularx}
\usepackage{gensymb}
\usepackage{todonotes}
\usepackage{array,multirow,graphicx}
\usepackage{diagbox}
\usepackage{booktabs}
\usepackage{siunitx}
\usepackage{tikz}
\usepackage{makecell}
\usepackage{multirow}
\usepackage[normalem]{ulem}
\usepackage{todonotes}
\usepackage{xspace}
\usepackage{array}
\usepackage{flushend}
\usepackage{cite}
\usepackage{float}
\usepackage{mathtools}
\usepackage{multirow}
\usepackage{booktabs}
\usepackage{hyperref}
\usepackage{cleveref}
\hypersetup{
    colorlinks=true,
    urlcolor=blue
}
\usepackage{balance}
\usepackage{fancyhdr}
\SetKw{Continue}{continue}
\SetKw{Break}{break}

\DeclareMathAlphabet{\mathcal}{OMS}{cmsy}{m}{n}

\author{Xing Li \quad\quad Manuel Baum \quad\quad Oliver Brock 
\thanks{All authors are with the Robotics and Biology Laboratory at Technische Universität Berlin and the Research Cluster of Excellence "Science of Intelligence". This work was supported by the Deutsche Forschungsgemeinschaft (DFG, German Research Foundation) under Germany’s Excellence Strategy – EXC 2002/1 “Science of Intelligence” – project number 390523135.}}

  \fancypagestyle{IEEECopyright}{
  \lfoot{\copyright2022 IEEE. Submitted to 2023 IEEE International Conference on Robotics and Automation (ICRA)}}

\begin{document}

\maketitle
\thispagestyle{IEEECopyright}
\pagestyle{empty}

\begin{abstract}

We introduce a Learning from Demonstration (LfD) approach for contact-rich manipulation tasks with articulated mechanisms. The extracted policy from a single human demonstration generalizes to different mechanisms of the same type and is robust against environmental variations. The key to achieving such generalization and robustness from a single human demonstration is to autonomously augment the initial demonstration to gather additional information through purposefully interacting with the environment. Our real-world experiments on complex mechanisms with multi-DOF demonstrate that our approach can reliably accomplish the task in a changing environment. Videos are available at the: \href{https://sites.google.com/view/rbosalfdec/home}{project page}.

\end{abstract}

\section{INTRODUCTION}


We introduce a Learning from Demonstration (LfD) approach that allows robots to robustly manipulate mechanisms in a changing environment based on a \emph{single} human demonstration. A single demonstration often does not contain all the necessary information for obtaining a general and robust policy. But instead of demanding additional demonstrations, we propose that robots augment the initial demonstration by additional information gathered autonomously.

\begin{figure}
	\centering
	\includegraphics[width=1\linewidth]{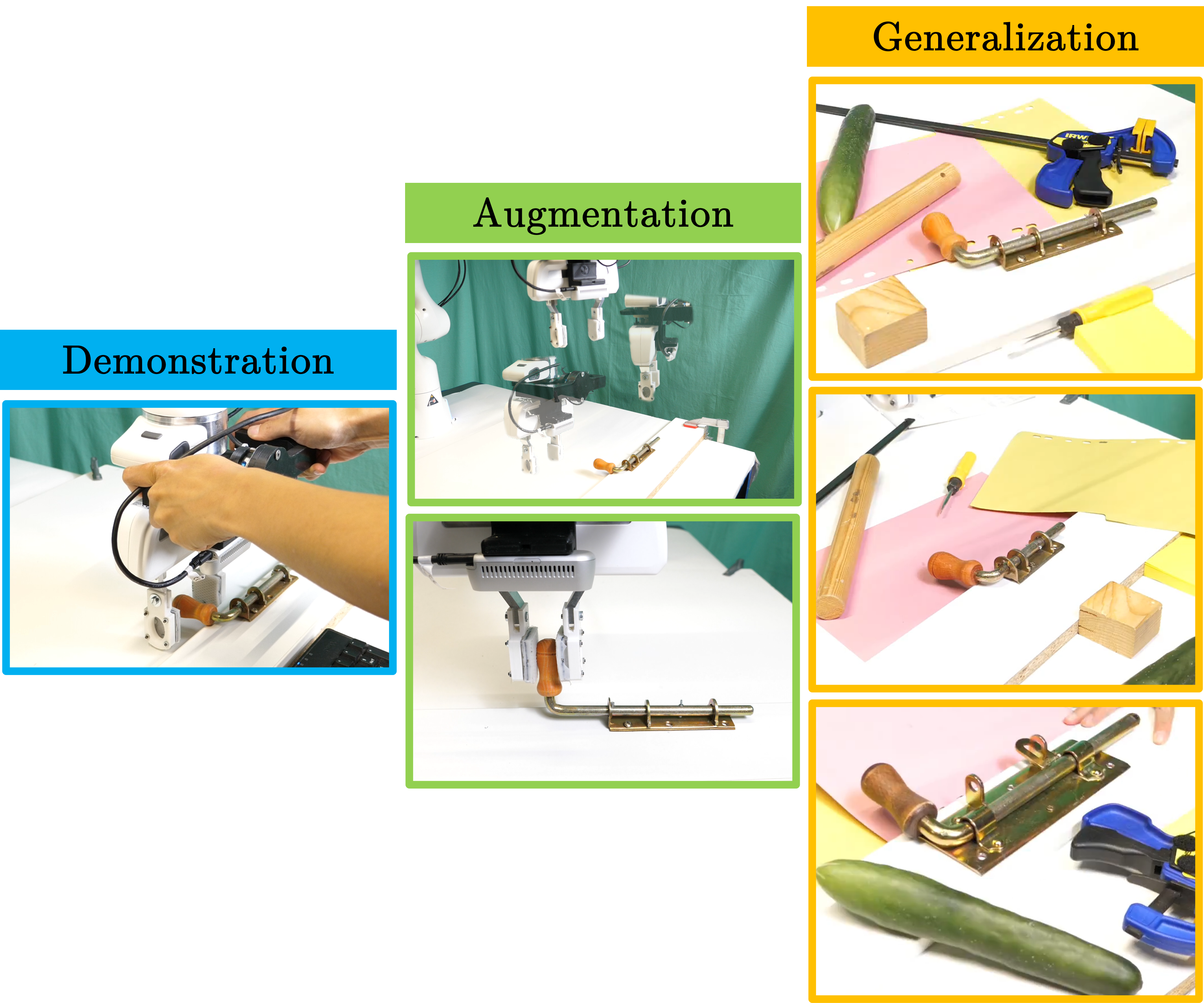}%
	\caption{This robot reliably opens three different locks based on a single demonstration of one lock. The extracted policy is also robust against environmental changes e.g. visual distractors. We achieve such substantial generalization and robustness by autonomously augmenting the initial demonstration to expose the complete information.}
	\label{fig:titlebild}
\end{figure}

To achieve general and robust LfD, robots should not precisely imitate the demonstrated actions but identify the constraints the robot must follow to solve the task. In mechanical manipulation tasks, these constraints are mainly vision- and contact-based. 
For example, visual constraints can guide a robot's end-effector to grasp a latch (Fig.~\ref{fig:titlebild}), and contact-based constraints can guide its motion to open that latch.

Policies are robust and generalizable when they properly utilize such constraints. Robustness is enabled because a focus on the right constraints helps reject distractors. Generalization is enabled because similar tasks often share similar structures, imposed by their visual and contact-based constraints.

The challenge is to identify such constraints from data, and often a single human demonstration is not sufficient. Robots need to gather additional information to identify task-relevant constraints.

Paradoxically, humans provide bad demonstrations because they have a way richer perceptual repertoire and prior knowledge than robots. Thus, they tend to exploit information in their demonstrations that robots cannot utilize. E.g. when humans are familiar with manipulated objects, their demonstrations might feature much less interaction with the environment~\cite{suomalainen2021survey}. But, guided by such an initial demonstration, robots can autonomously augment their data about task-relevant visual and contact-based constraints. They can explore different visual perspectives to make their behavior more robust, different kinematic constraints to learn contact-rich manipulation strategies.

We demonstrate that a policy extracted from an augmented demonstration with a mechanical lock (see Fig.~\ref{fig:titlebild}) generalizes across different locks of the same type and yields robust behaviors against environmental changes, achieving a $90\%$ success rate. We also demonstrate that a \emph{single} human demonstration with a drawer suffices to open another drawer, achieving $100\%$ success rates. Our key contribution is to present this LfD approach, which incorporates augmentation to extract general and robust policies for manipulating mechanisms. 

\section{RELATED WORK}

Our LfD approach aims to learn a policy that exploits constraints of the environment from an augmented demonstration. Therefore, we discuss these three topics: Learning from Demonstration, Environmental Constraint Exploitation, and Augmentation to motivate our integrated approach. 

\subsection{Learning from Demonstration}

Learning from Demonstration has been shown as a promising way to teach robots manipulation skills~\cite{billard2008robot}. A big portion of LfD approaches focus on precisely imitating demonstrated low-level actions (e.g. Behavior Cloning)~\cite{pomerleau1988alvinn, zhang2018deep, Rahmatizadeh2018} or learning movement primitives which encode motion characteristics of demonstrations~\cite{pastor2009learning, paraschos2013probabilistic, manschitz2015learning}. Policies obtained from Behavior Cloning are often trained end-to-end, thus data-hungry. On the other hand, learning movement primitives assumes the environment is well-structured e.g. object poses are known. In this work, we augment the initial demonstration rather than asking for more demonstrations, achieving efficient LfD. Additionally, the extracted policy exploits rich feedback provided by the environment, improving robustness in unstructured environments.

\subsection{Environmental Constraint Exploitation}

Humans extensively exploit useful features provided by the environment during manipulation.These features are referred to as Environmental Constraints (EC)~\cite{eppner_exploitation_2015}. ECs provide rich and helpful feedback, which greatly simplifies perception and control, leading to robust robot behaviors~\cite{castano1994visual, lozano1984automatic, eppner_planning_2015, Bhatt-RSS-21, baum2022icra}. Unlike these approaches relying on highly specific controllers, our approach utilizes general controller templates that exploit ECs and instantiate them for specific instances with a human demonstration. Moreover, our approach mitigates the impact of less informative demonstrations through efficient augmentation, ensuring proper instantiation. 

\subsection{Augmenting Demonstration}

The explorative properties of RL can be leveraged to offer additional interaction with the environment to augment human demonstrations, resulting in robust policies~\cite{vecerik2017leveraging, nair2018overcoming, rajeswaran2017learning, zhu2018reinforcement, su2020reinforcement}. However, the tasks we tackle involve extensive physical contact in a chaotic real-world environment, which is difficult to be simulated. Moreover, small position errors could generate a large and dangerous reaction force when manipulating mechanisms. Hence, using RL to augment demonstrations is not suited for our problem. 

In contrast to the RL-based LfD methods,~\cite{fu2019active} and~\cite{johns2021coarse} introduce approaches that augment a demonstration to obtain self-supervised datasets for learning vision-based skills e.g. grasping. However, these approaches are limited in scenarios with significant environmental changes which are not covered by the training datasets. Therefore, we extend these approaches by separating general information from scenario-specific information before learning, facilitating generalization. Additionally, apart from vision-based skills, our approach extracts a general and robust policy for contact-rich tasks by exploiting benefits provided by environmental constraints. 

\section{Environmental Constraints For Manipulation of Mechanisms}

Our goal is to, from a human demonstration, derive a policy that builds on task-relevant constraints for manipulating mechanisms. We call these constraints Environmental Constraints~\cite{eppner_exploitation_2015}. In this section, we explain two types of ECs that are essential for manipulating mechanisms, including grasping and opening. 

Many mechanical manipulation tasks require robots to act either in contact with the environment, e.g. to manipulate DoF of articulated objects, or without contact, e.g. moving in free-space to establish grasps. Those aspects of a task that are not contactful can be defined by \emph{visual} constraints that capture the relation between the robot and its environment visually. Such constraints capture, among other things, what a promising grasping configuration looks like. We call such constraints vision-based ECs. Complementarily, contact-based ECs capture those constraints that are imposed by the kinematics of manipulated mechanisms. They capture the spatial directions in which the robot can move an articulated mechanism, but also the directions in which it can excert forces.

Both kinds of constraints can be exploited using control methods.
A proven solution to exploit the guidance of vision-based ECs is visual servoing~\cite{hutchinson1996tutorial}, which we also use in our method. To instantiate this controller template for specific tasks, we have to derive vision-based, such as a constellation  of visual features that define a grasp pose. To exploit contact-based ECs, we use adaptive compliant controllers developed in our previous work~\cite{li23ral}. To execute this controller for a particular object, we need properties of contact-based ECs, which are motion and force directions. 

To summarize, manipulation of mechanisms can be represented as a sequence of (vision- and contact-based) EC exploitations, similar to grasping and in-hand manipulation tasks~\cite{eppner_exploitation_2015, Bhatt-RSS-21}. Therefore, we aim to develop a novel LfD that extracts a sequence of EC exploitations from an \emph{augmented} demonstration for manipulating mechanisms.

The following sections will present the idea of identifying vision- and contact-based ECs from a human demonstration. In addition, we will explain why information from human demonstration is insufficient to acquire EC exploitations and how to fill in the missing information through augmentation.

\section{Augmenting Demonstration for Vision-Based Environmental Constraints}
\label{sec:method:vec}

In this section, we concentrate on the grasping problem. Concretely, we first explain our general controller template to exploit vision-based ECs for grasping, then discuss the details of instantiating this controller template through learning from a demonstration with augmentation. 
 
\subsection{Exploitation of Vision-Based ECs}
\label{sec:method:vec:exploitation} 

In the context of grasping, a grasp pose acts as a vision-based EC to restrict the position of a robot's end-effector. Our approach assumes an eye-in-hand setting, where a camera is mounted to the robot's end-effector. We define the relative pose between the end-effector $E$ and the target grasp pose $G$ as $_{\mathrm{E}}^{\mathrm{G}}T$. To achieve a grasp pose, we use a pose-based visual servoing controller~\cite{haviland2022holistic}, which modulates the difference between the desired grasp pose and the end-effector pose into velocity signals to control the robot. We terminate a servoing controller when the grasp pose is reached, or a collision is detected. Now, the only aspect to be learned from demonstration is an estimator for a grasp pose of an instance. 

\subsection{Augmentation for Vision-Based ECs}
\label{sec:method:vec:augmentation}

A human demonstration does not contain enough information about grasping pose i.e. vision-based ECs. Consider a task that humans teach a robot with eye-in-hand configuration how to grasp an object. When they provide a kinesthetic demonstration of grasping, they often straightforwardly move the end-effector towards the target object, follwing a near linear path without much exploration. As a result, this exploitive demonstration could not provide visual information about the target object from different perspectives. One reason for human demonstration being too exploitive is that humans perform demonstrations that are supported by a different, way richer perceptual repertoire than robots. Task-relevant visual information can be easily interpreted by humans, but not by robots! We now introduce augmentation to bridge the gap of perception repertoire between humans and robots. 

Our augmentation approach is inspired by~\cite{fu2019active} and~\cite{johns2021coarse}. The principle of the idea is this: assuming the object's pose is fixed during training, we could directly assign the last demonstrated end-effector pose $_{\mathrm{B}}^{\mathrm{E}}T$ as the grasp pose $_{\mathrm{G}}^{\mathrm{B}}T$, where B is the robot base frame. By leveraging this assumption, we could actively move the robot around the grasp pose and collect images together with the known grasp pose $_{\mathrm{E}}^{\mathrm{G}}T$ as labels. 

Unlike the previous approaches presented in~\cite{fu2019active} and~\cite{johns2021coarse} directly regress images to the grasp pose $_{\mathrm{E}}^{\mathrm{G}}T$, we believe that it is better to first separate the task-relevant visual information with the information specific to scenarios. We assume that the task-relevant visual features are surrounded by the demonstrated grasp pose, while the outside of the area e.g. background, is task-irrelevant and scenario-specific information. Following this idea, we divide the grasp pose estimation problem into two sequential components: 1) an object-agonistic detector; 2) a pose estimator based on detected results. 

For estimating a grasp pose, our network first detects the area of the grasping location as a bounding box. The center point $\mathrm{c}=\left(c_{x}, c_{y}\right)$ of the bounding box is the grasp position in the camera frame. We can then calculate the grasp position $\mathrm{p}=\left(p_{x}, p_{y}, p_{z}\right)$ using a depth camera. Secondly, our network predicts the grasp orientation only using visual features cropped by the detected bounding box. We consider the tasks where a mechanism is mounted on the surface, and the end-effector is constrained to be perpendicular to the surface. Hence, our network only has to predict the yaw angle $\Psi \in \mathbb{R}$ in the end-effector frame. We implement this network by extending the framework of Faster-RCNN with a ResNet-50-FPN backbone in Pytorch~\cite{ren2015faster}. In addition to a bounding box regressor and a classifier, we add a layer to estimate $\Psi$ for each potential object in the image.

\begin{figure}
\centering
  \includegraphics[width=0.8\linewidth]{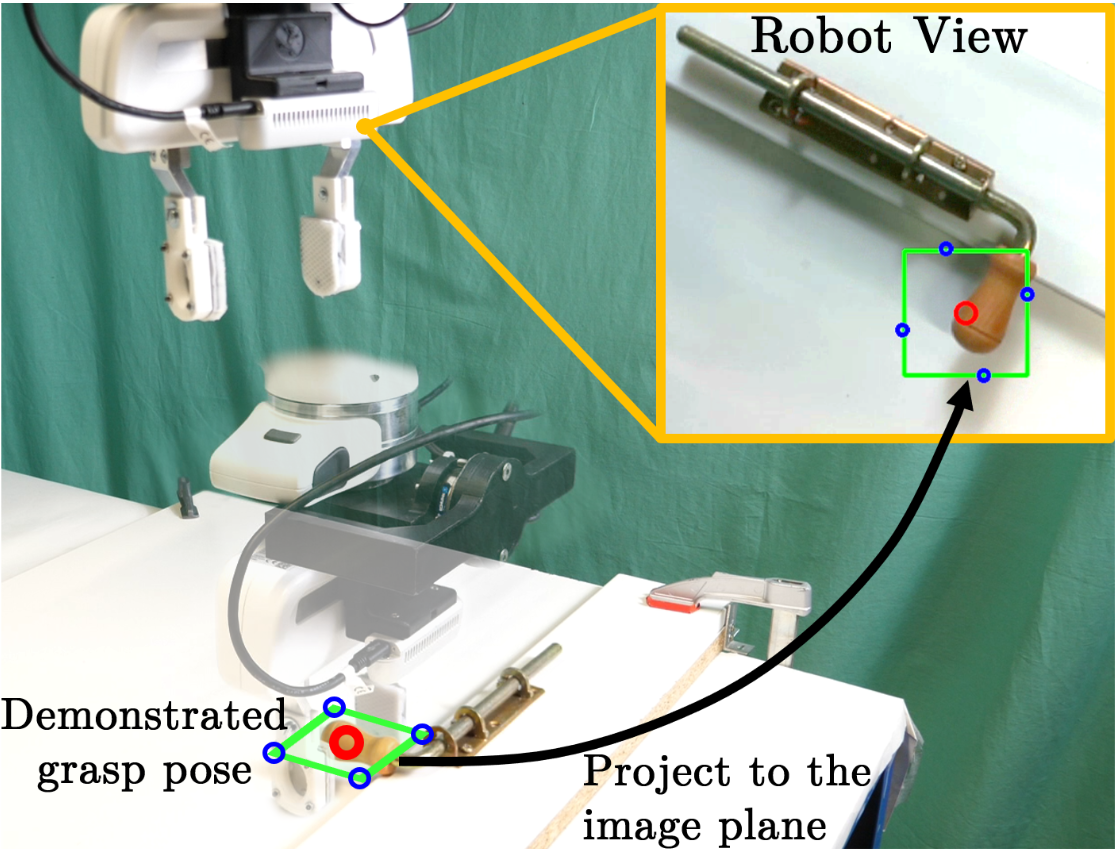}%
  \caption{This figure shows how to generate training data. We crop an image based on the known grasp position and back-project it to the image plane to create a bounding box.}
   \label{fig:generateVisualData}
\end{figure}

Training this network requires bounding boxes as labels. Fig~\ref{fig:generateVisualData} shows the process of generating training data. First, we generate a green square in the x-y plane of the end-effector frame. The red center point of the square is the demonstrated grasp position. The length of the square is based on the object's width obtained from the demonstrated grasping action. Then, the four blue vertices are back-projected to the image plane to create a bounding box as the training label. We use the bounding box and the rotation angle $\Psi$ derived from the known $_{\mathrm{E}}^{\mathrm{G}}T$ to train our network. 

One of the key advantages of our network is that the object detection pipeline enables us to recognize the scenario where no target object is visible. The robot then could trigger a predefined behavior in response to this scenario, such as searching the object. Besides, our approach concentrates on the most relevant subarea of the image restricted by the bounding box to infer the grasp pose, improving robustness against environmental changes. We demonstrate these benefits in the experiments.  

\section{Augmenting Demonstration for Contact-Based Environmental Constraints}
\label{sec:method:cec}

We now introduce how to open a mechanism through physical contact after a grasp was performed. Again, we will first introduce the general controller template for exploitation of contact-based ECs in \Cref{sec:method:cec:exploitation} and explain the details of augmentation in \Cref{sec:method:cec:augmentation}.

\subsection{Exploitation of Contact-Based ECs}
\label{sec:method:cec:exploitation}

We can exploit contact-based ECs through deliberate contact with the environment, where compliance is essential to ensure safe interaction. In this work, we use the adaptive compliant controller, developed in our previous work~\cite{li23ral}, as the fundamental controller template of the policy. 

This controller simultaneously estimates an admissible motion direction, and compliantly controls the robot's velocity in this direction. With this controller, the robot could follow the guidance provided by contact-based ECs. To operate complex mechanisms with multi-DOF, we sequence a set of controllers with contact-changing events. While following a constraint, the activated controller also exerts a force to maintain contact with the mechanism, which facilitates recognizing contact-changing events (gain new contacts or lost contacts) for transitions. 

To instantiate an adaptive compliant controller, we need a pair of motion $\hat{m}$ and force $\hat{f}$ directions. In our previous work, we extract these motion and force directions from a demonstration. Concretely, we segment the demonstrated position and force trajectory $Tr$ and calculate a set of vector pairs of motion and force directions $Tr = \left\{ \left(\hat{m}_1, \hat{f}_1\right), ...,\left(\hat{m}_k, \hat{f}_k\right) \right\}$ with $k$ segments. These vector pairs are then used to instantiate a sequence of adaptive compliant controllers to reproduce the demonstrated opening action.

\subsection{Augmentation for Contact-Based ECs}
\label{sec:method:cec:augmentation}

\begin{figure}
\centering
  \includegraphics[width=1\linewidth]{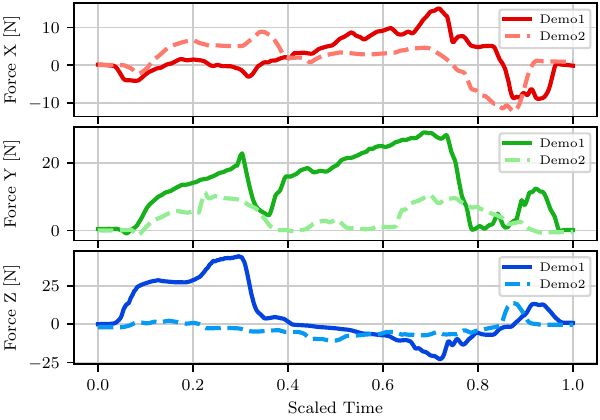}%
  \caption{Force profiles of two \emph{successful} demonstrations on opening the lock in Fig.~\ref{fig:titlebild}. The dissimilarity of force profiles indicates that human demonstrations cannot guarantee reliable and constant valid force information.}
   \label{fig:ForceProfiles}
\end{figure}

So far, we have explained that motion and force information (i.e. directions) is required to reproduce the contact-based EC exploitations for opening a mechanism. However, human demonstrations might not contain enough force information for learning compliant motions~\cite{kalakrishnan2011learning, racca2016learning}. If the demonstrator is familiar with the manipulated objects, the provided demonstration might contain much less interaction with the environment~\cite{suomalainen2021survey}. As an example, Fig.~\ref{fig:ForceProfiles} shows two kinesthetic demonstrations of successfully opening a lock in Fig.~\ref{fig:titlebild}. The dissimilarity of the force profiles indicates that human demonstrations might not yield reliable force information. But reliable force information can be gathered, as we propose, by augmenting the demonstration. 

\begin{figure}
	\centering
	\includegraphics[width=0.8\linewidth]{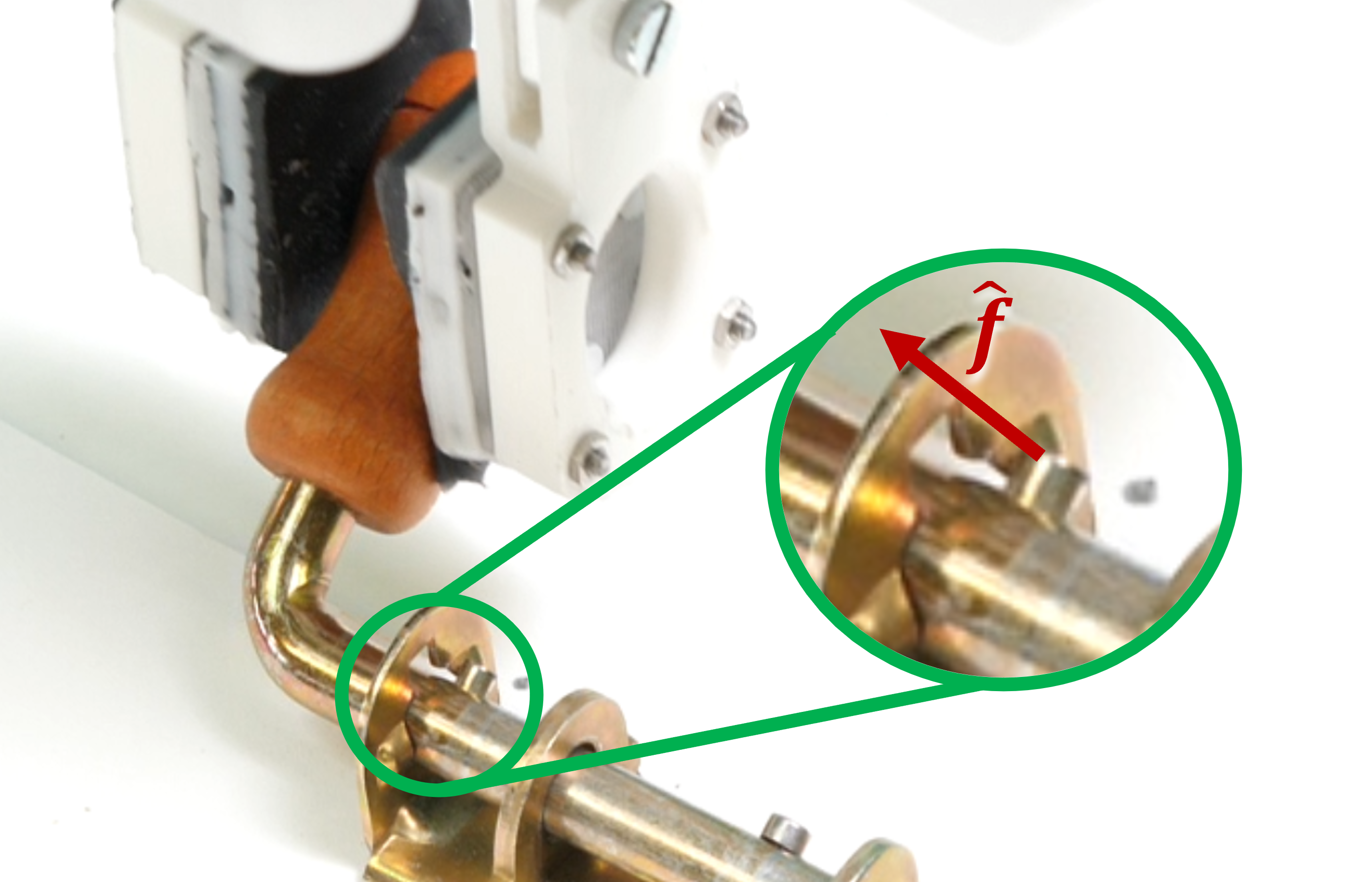}%
	\caption{A robot is lifting a lock to pass the small pin through the narrow slot. Based on the force direction obstained through augmentation, the robot applies force to deliberatly maintain contact against the contraints. It allows the robot to eailsiy pass through the narrow slot.}
	\label{fig:CECAugmentation}
\end{figure}

The goal of augmentation is to fill in the necessary force directions. Different to force directions, the demonstrated motion directions are reliable as the demonstrator should follow the dications of the contact-based ECs. Our key idea is to hypothesize a set of force directions based on these reliable motion directions. This idea stems from the observation of the narrow passage problem: maintaining contact with the future motion direction through the passage facilities the robot to pass through the narrow passage, as shown in Fig.~\ref{fig:CECAugmentation}. In addition to future motion direction, it is also beneficial in some cases to maintain contact using the past motion direction. Having to maintain this contact reduces possible DOFs, thus stabilizing the motion. Besides, we propose a third force direction hypothesis: direction of gravity, as objects usually have support against gravity. In this way, given a motion direction $\hat{m}_i$ in the $i$ segment, the force hypotheses are $f_h = \left(\hat{m}_{i+1}, \hat{m}_{i-1}, \hat{f}_g\right)$.

The Algorithm~\ref{alg:SACEC} presents how to evaluate hypothesized force directions. The goal of having force directions is to maintain contact. Thereby, our evaluation criterion is based on the observed movement. Concretely, the robot applies a force in the hypothesized force direction. This force hypothesis is valid if no movement is observed. When we found the valid force direction $\hat{f}_i$ for the motion direction $\hat{m}_i$ in the $i$ segment of the demonstration, the robot must move to the start position $p_{i+1} \in \mathbb{R}^{3}$ of the next segment $i+1$ to evaluate for the motion direction $\hat{m}_{i+1}$. It is worth mentioning that replaying a demonstrated trajectory can not always arrive at the position $p_{i+1}$ due to the properties of contact-rich manipulation such as friction. Therefore, we instantiate an adaptive compliant controller using the $\left(\hat{m}_i, \hat{f}_i\right)$ to move the robot to $p_{i+1}$. In this way, we complete force directions through augmentation instead of using the unreliable demonstrated force information, guaranteeing the performance of our approach. 
 
\begin{algorithm}
\SetAlgoLined
\DontPrintSemicolon
\KwIn{A segmented demonstration $\left\{ \left(p_1, \hat{m}_1\right), ..., \left(p_k, \hat{m}_k\right) \right\}$ where $p_i$ is the start position of end-effector in the $i$ segment.}\KwOut{A sequence of motion and force directions $\left\{\left(\hat{m}_1, \hat{f}_1\right), ..., \left(\hat{m}_k, \hat{f}_k\right)\right\}$}
 
\For{\texttt{$i = 1, ..., k$}}
{
	Generate force hypotheses $\hat{f}_h = \left(\hat{m}_{i+1}, \hat{m}_{i-1}, \hat{f}_g\right)$ for $\hat{m}_i$
	
	\ForEach{$\hat{f} \in \hat{f}_h$}{
		The robot applies a force into $\hat{f}$ direction
		
        \If {No movement is observed}
        {
			Save $\hat{f}$ as the valid force direction for $\hat{m}_i$
			
			\Break
		}
	}

	Instantiate adaptive compliant controller with $\left(\hat{m}_i, \hat{f}\right)$
	
	Move the robot to the next starting position $p_{i+1}$ using the adaptive compliant controller
}	
\caption{Augmentation for Force Directions}
\label{alg:SACEC}
\end{algorithm}

\section{Experiments And Results}
We test the capabilities of our approach in real-world experiments on contact-rich manipulation tasks with various articulated mechanisms. The experiments are conducted on a 7-DOF Panda arm with a 1-DOF gripper. We attached a RealSense Depth Camera D435i to the end-effector and an ATI-mini 40 Force/Torque sensor to the wrist. The quantitative results prove that our approach is capable of inferring a policy to robustly grasp and open mechanisms based on a single human demonstration with augmentation. Videos for experiments are available at the: \href{https://sites.google.com/view/rbosalfdec/home}{project page}.

\subsection{Opening Locks}
\begin{figure}
	\centering
	\includegraphics[width=0.9\linewidth]{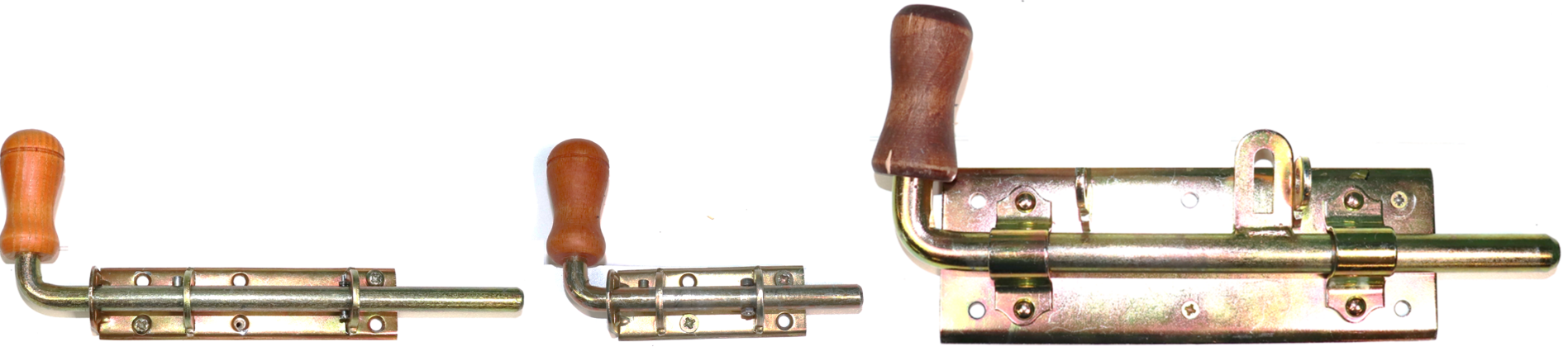}%
	\caption{This figure shows lock 1, lock 2 and lock 3 from left to right, which are used in the experiment.}
	\label{fig:alllocks}
\end{figure}

In this experiment, our goal is to test the generalization and robustness of the proposed approach for lock opening tasks. Given a demonstration with lock 1 (see Fig.~\ref{fig:alllocks}), our approach augments this demonstration to obtain an EC-based policy. We then test the generalization of the policy on all three locks. As the lock opening task involves two sub-tasks: grasping and opening, we divided this experiment into three parts. In the first two parts, we evaluate the grasping and opening sub-tasks separately. Lastly, we demonstrate the overall efficacy of our approach by showing that it allows a robot to grasp and open (full-task) a lock in a dynamic and cluttered environment, just from a single human demonstration with augmentation.

\begin{table*}
	\footnotesize
	\centering
	\begin{tabular}{ccccccccccccc} 
		\toprule
		\multirow{2}{*} & 	  \multicolumn{4}{c}{Grasp} & \multicolumn{4}{c}{Open} & \multicolumn{4}{c}{Full-Task}  \\
					  		  \cmidrule(lr){2-5} 			 \cmidrule(lr){6-9} 		   \cmidrule(lr){10-13}
		Method                & Lock 1 & Lock 2 & Lock 3 & Avg.    & Lock 1 & Lock 2 & Lock 3 & Avg.   & Lock 1 & Lock 2 & Lock 3 & Avg.      \\
		\midrule
		 w/o augmentation     & 0\%    & 0\%    & 0\% & 0\%       & 0\%    & 30\%   & 20\% & 17\%     & -      & -      & -  & -          \\
		 w/ augmentation      & 100\%  & 100\%   & 100\%  & 100\%     & 100\%  & 100\%  & 100\%  & 100\%    & 100\%  & 100\%   & 70\% & 90\%         \\
		\bottomrule
	\end{tabular}
	\caption{This table compares the success rate of our approach with and without augmentation for grasping, opening, and combining as a complete task for three locks. Augmentation significantly improves the success rate.}
	\label{table:LockResults}
\end{table*}
\subsubsection{Grasping}

Based on a demonstration of grasping, the robot augments this demonstration by following a funnel-like trajectory and collects additional training data. We collect $2498$ images in $3$ min with augmentation for training. As a comparison, we train the same model only on the demonstration data, which contains $50$ images without rotation. During the test, a lock is randomly placed on the table. Additionally, we purposefully change the pose of the lock when the robot approaches the lock. We do so to challenge our approach in a dynamic environment. We run 10 grasping trails for each lock. A grasp trial is successful if the robot grasps the knob and lifts it. 

\begin{figure}
	\begin{minipage}[t]{0.49\linewidth}
	\includegraphics[width=\linewidth]{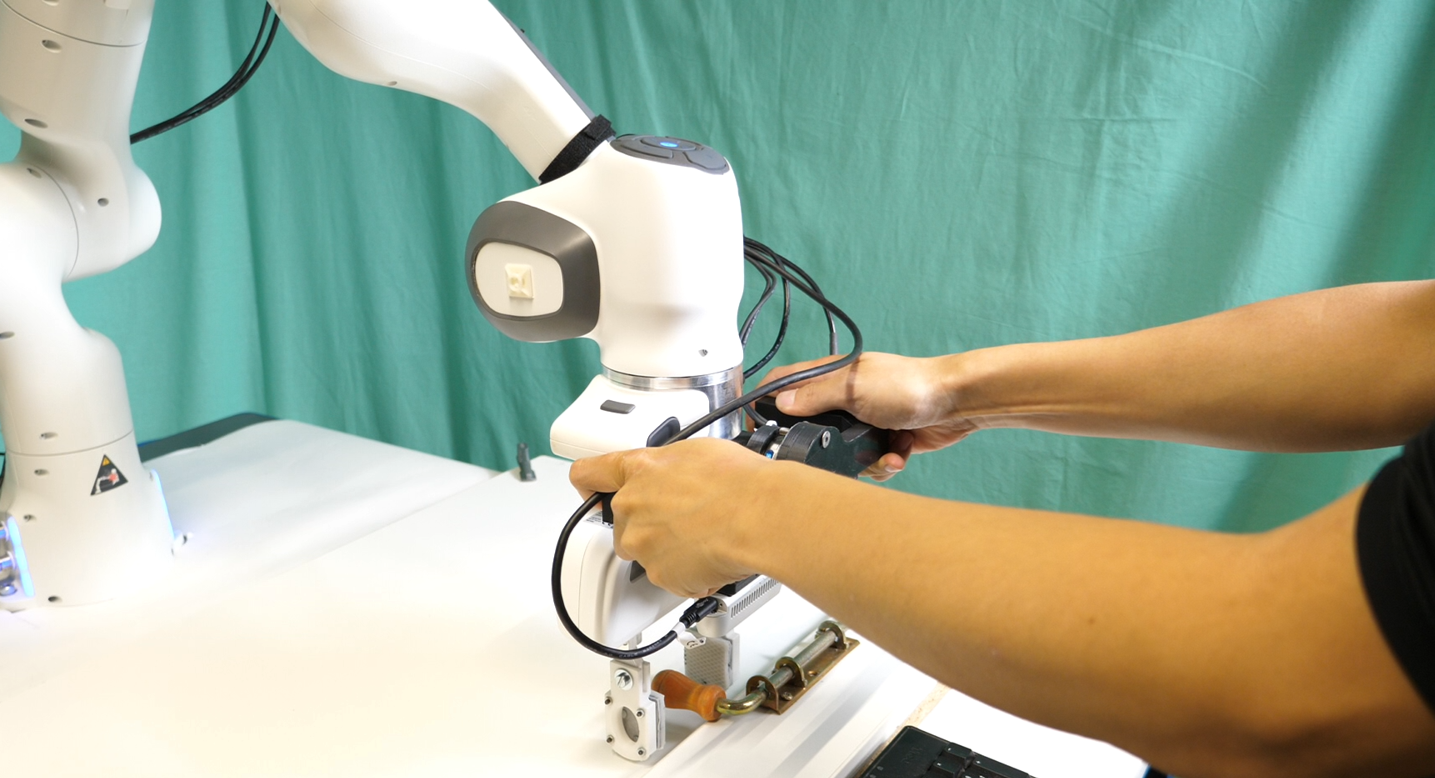}
\end{minipage}
\begin{minipage}[t]{0.49\linewidth}
	\includegraphics[width=\linewidth]{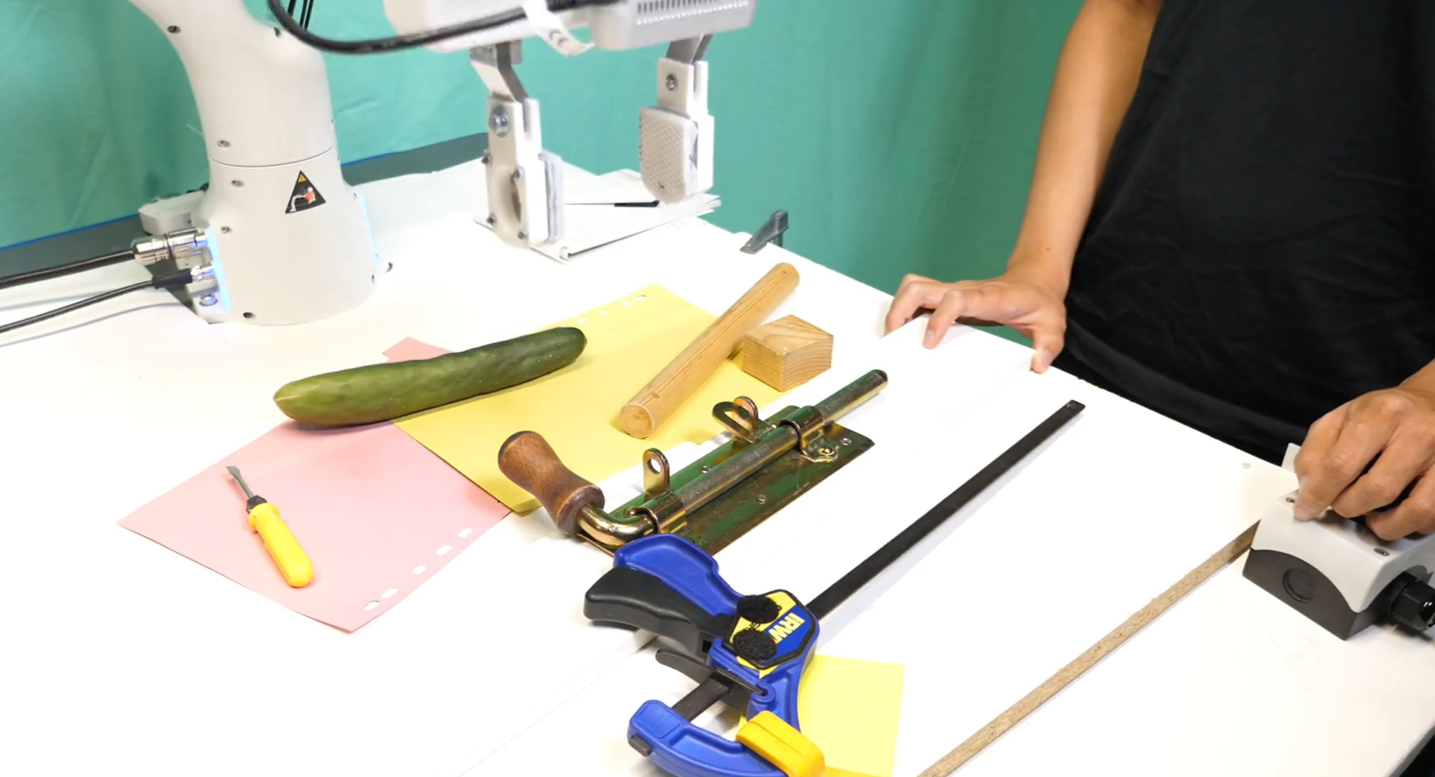}
\end{minipage}
\caption{We provide and augment a demonstration on grasping and opening a lock in a clean environment (left). The extracted policy generalizes to a different lock in a dynamic and cluttered environment (right).}
     \label{fig:VECexperiment}
\end{figure}

Table~\ref{table:LockResults} presents the success rate of grasping. Our approach with augmentation succeeds 10/10 times on the same lock used in the demonstration, highlighting the robustness of the approach in dynamic environments achieved by constantly exploiting useful feedback provided by the vision-based EC. In contrast, the approach without augmentation fails to grasp the demonstrated lock. Most failure cases happen when humans rotate the lock, and the network cannot predict the orientation correctly, as the demonstrated data does not contain data for rotated images. We attribute this to the fact that the demonstration is not informative enough to infer a robust EC-based policy for grasping. These results support our argument that the gap in vision-based manipulation repertoire between humans and robots can be compensated by augmentation, leading to a robust grasping behavior. 

Note that the extracted policy on lock 1 can reliably grasp other locks with $100\%$ success rates. This superior generalization ability is achieved by using ECs as the representation, which excludes task-irrelevant visual features e.g. the lock's backbone for the learning system. 

\subsubsection{Opening}

In the second part of the experiment, we validate our approach to open locks after a grasp was established. We first provide a \emph{successful} demonstration on opening lock 1 with low contact forces between gripper and environment, as shown in Fig.~\ref{fig:ForceProfiles}. Our approach then augments this demonstration to obtain force directions. Based on the demonstrated motion directions and augmented force directions, our approach instantiates a sequence of adaptive compliant controllers for opening all locks. To analyze the impact of augmentation, we use the method developed in~\cite{li23ral} as a baseline, which extracts all necessary parameters directly from a demonstration and does not use augmentation. 

We run 10 trials per lock for each method. As clearly seen in Table~\ref{table:LockResults}, the approach using force information from demonstration performs poorly, only succeeding 5 out of 30 times i.e. $17\%$ for all locks. Most failure trials occur when the robot attempts to pass the pin through the narrow slot (see Fig.~\ref{fig:CECAugmentation}). This failure occurs because it is difficult for robots to deal with the narrow passage problem without maintaining contact orthogonal to the passage. It proves that demonstration with less interaction with the environment cannot guarantee reliable force information to exploit contact. Conversely, our augmentation significantly improves the success rate to $100\%$ for all locks. This improvement is due to augmentation, which allows the robot to extract constraints and force directions by interaction, filling in missing force directions for contact-based ECs. 

\subsubsection{Full-Task Execution}

After we examined grasping and opening individually, we then test the method's performance to grasp and then open locks in the same experiment. We challenge the robustness of our approach in a dynamic and cluttered environment. Concretely, we place distractors on the table and change the position of the lock when the robot is approaching. We run 10 trials per lock and calculate the success rate. The result is depicted in Table~\ref{table:LockResults}. Our approach achieves $100\%$ for locks 1 and 2 even in this highly uncertain environment. Note that the success rate of lock 3 decreases to $70\%$ due to the false-negative detection for the knob, which exposes the limitation of the generalization ability of our approach for grasping. We will discuss this limitation in~\Cref{sec:limitations}. Moreover, we show that our approach could recognize the scenario where the lock is not visible and triggers a search behavior (see supplementary video), contributing to robustness. These experimental results provide strong evidence that our approach can extract general and robust policy from a single human demonstration with augmentation.

\subsection{Opening Drawers}

\begin{figure}
	\begin{minipage}[t]{0.49\linewidth}
	\includegraphics[width=\linewidth]{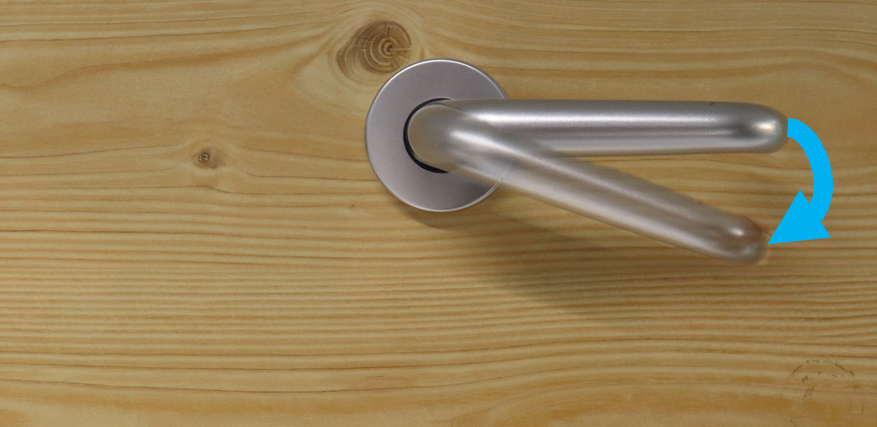}
\end{minipage}
\begin{minipage}[t]{0.49\linewidth}
	\includegraphics[width=\linewidth]{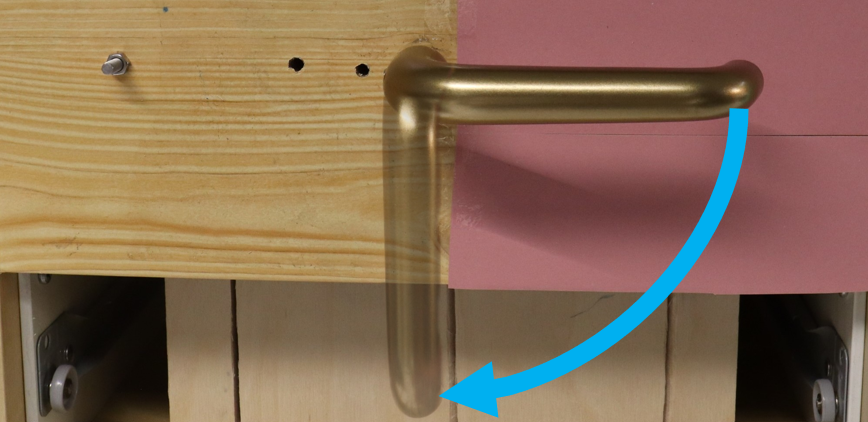}
\end{minipage}
\caption{The extracted policy on the siliver drawer handle (left) generalizes to the gold one (right) with different rotation angle for opening and background.}
     \label{fig:twoDrawers}
\end{figure}

We also conduct experiments for learning to grasp and open a drawer with a revolute handle, including transfer of the policy to another drawer with different kinematics and appearance, as shown in Fig.~\ref{fig:twoDrawers}. Based on a human demonstration on a drawer, our approach derives a grasping policy based on an augmented dataset collected in 3 minutes of augmentation time. When opening the drawer, our approach first exploits the contact-based EC provided by the revolute joint of the handle. At the same time, it maintains contact in the direction of pulling using the augmented force direction. Maintaining this contact generates a contact-changing event to transit to the subsequent exploitation of contact-based EC derived from a prismatic joint. 

To compute the success rate, we run 10 trials to open this drawer in a dynamic environment (i.e., human disturbances) with changes in the background. Our approach succeeds 10/10 times. Furthermore, the success rate remains $100\%$ even for a different drawer (see Fig.~\ref{fig:twoDrawers}). These results confirm that our approach can not only operate various mechanisms robustly -- from a single augmented demonstration -- but also that it yields behavior that generalizes to unseen object instances. 

\section{Limitations}
\label{sec:limitations}

The limitations of our proposed LfD approach with augmentation are twofold. First, as mentioned before, the performance decreases for lock 3 shows that the extracted policy for grasping cannot reliably generalize to objects with distinct visual features. This limitation results in solely using the color information to represent vision-based ECs. We are working on integrating 3D geometric information into our approach to achieving better generalization. As a second limitation, our applications are restricted to manipulation of mechanisms, which assume the target is fully constrained in the environment. In this context, our approach can augment vision- and contact-based ECs that physically exist in the environment but not task constraints e.g. reference frames~\cite{ureche2015task} or joint dependency of a lockbox~\cite{baum2017opening}. Our next research goal is to extend our approach by incorporating human physical corrections to reveal the information about task constraints~\cite{laskey2017dart, tahara22icra}. We believe this approach will allow us to apply the idea of using ECs in LfD to a wide range of robotic manipulation tasks. 

\section{CONCLUSIONS}

We present a LfD approach that leverages the potential provided by the robots' embodiment to augment a demonstration autonomously. Our real-world experiments demonstrate that our approach allows a robot to learn from a single human demonstration and robustly manipulate various mechanisms by exploiting environmental constraints. Moreover, the extracted policy can be transferred to unseen mechanisms that vary in size and appearance but share similar vision- and contact-based environmental constraints. The insight we found in this paper is that demonstrations are not always informative enough for a robot to extract EC-based manipulation policies. Our approach exposes the complete information for ECs by augmenting a human demonstration. Last but not least, this research opens an intriguing research problem: how to fill in the missing information that augmentation cannot deal with.

\balance
\addtolength{\textheight}{-6cm}   

\bibliographystyle{IEEEtran} 
\bibliography{references}

\end{document}